\documentclass[runningheads]{llncs}

\PassOptionsToPackage{hyphens}{url} 
\usepackage[hidelinks]{hyperref} 

\usepackage[T1]{fontenc} 
\usepackage[utf8]{inputenc} 

\usepackage[main=english, vietnamese]{babel} 
\newenvironment{vietnamese*}
{\begin{otherlanguage*}{vietnamese}}
	{\end{otherlanguage*}} 

\usepackage[hang]{footmisc}  
\usepackage{enumitem}  
\usepackage{setspace}  

\usepackage{lipsum} 
\usepackage[disable]{todonotes} 

\usepackage{graphicx} 
\usepackage{subfigure}
\usepackage{tikz} 

\usepackage{booktabs} 
\usepackage{multirow} 
\usepackage{multicol} 
\usepackage{makecell} 
\usepackage{array, tabularx, boldline} 

\usepackage{amsmath} 
\usepackage{amssymb} 

\usepackage{amsthm}
\usepackage{amsfonts}
\usepackage{bm} 
\usepackage{mathtools} 
\usepackage{mathdots} 
\usepackage{xfrac} 
\usepackage{faktor} 
\usepackage{cancel} 

\theoremstyle{plain}  

\theoremstyle{definition}  
\newtheorem{defn}{Definition}

\theoremstyle{remark}  

\def\eqref#1{equation~\ref{#1}}

\def\1{\bm{1}}

\def\ve{{\bm{e}}}

\def\vh{{\bm{h}}}

\def\vr{{\bm{r}}}

\def\vt{{\bm{t}}}
\def\vu{{\bm{u}}}
\def\vv{{\bm{v}}}

\DeclareMathAlphabet{\mathsfit}{\encodingdefault}{\sfdefault}{m}{sl}
\SetMathAlphabet{\mathsfit}{bold}{\encodingdefault}{\sfdefault}{bx}{n}

\def\gE{{\mathcal{E}}}

\def\gS{{\mathcal{S}}}

\def\sR{{\mathbb{R}}}

\DeclareMathOperator*{\argmax}{arg\,max}

\begin{document}
%
%

\renewcommand\floatpagefraction{1.}
\renewcommand\topfraction{1.}
\renewcommand\bottomfraction{1.}
\renewcommand\textfraction{.0}
\setcounter{totalnumber}{50}
\setcounter{topnumber}{50}
\setcounter{bottomnumber}{50}


\title{Exploring Scholarly Data by Semantic Query \\on Knowledge Graph Embedding Space}
\titlerunning{Semantic Query on Knowledge Graph Embedding Space}

\author{Hung Nghiep Tran\inst{1} \and
Atsuhiro Takasu\inst{2}}
%
%
\institute{SOKENDAI (The Graduate University for Advanced Studies), Tokyo, Japan \and
National Institute of Informatics, Tokyo, Japan \\
\email{\{nghiepth, takasu\}@nii.ac.jp}}

\maketitle              

\makeatletter
\def\blfootnote{\gdef\@thefnmark{}\@footnotetext}
\makeatother
\blfootnote{\scriptsize The 23rd International Conference on Theory and Practice of Digital Libraries (TPDL2019).}

\begin{abstract}
The trends of open science have enabled several open scholarly datasets which include millions of papers and authors. Managing, exploring, and utilizing such large and complicated datasets effectively are challenging. In recent years, the knowledge graph has emerged as a universal data format for representing knowledge about heterogeneous entities and their relationships. The knowledge graph can be modeled by knowledge graph embedding methods, which represent entities and relations as embedding vectors in semantic space, then model the interactions between these embedding vectors. However, the semantic structures in the knowledge graph embedding space are not well-studied, thus knowledge graph embedding methods are usually only used for knowledge graph completion but not data representation and analysis. In this paper, we propose to analyze these semantic structures based on the well-studied word embedding space and use them to support data exploration. We also define the semantic queries, which are algebraic operations between the embedding vectors in the knowledge graph embedding space, to solve queries such as similarity and analogy between the entities on the original datasets. We then design a general framework for data exploration by semantic queries and discuss the solution to some traditional scholarly data exploration tasks. We also propose some new interesting tasks that can be solved based on the uncanny semantic structures of the embedding space.

\end{abstract}

\section{Introduction}
In recent years, digital libraries have moved towards open science and open access with several large scholarly datasets being constructed. Most popular datasets include millions of papers, authors, venues, and other information. Their large size and heterogeneous contents make it very challenging to effectively manage, explore, and utilize these datasets. The knowledge graph has emerged as a universal data format for representing knowledge about entities and their relationships in such complicated data. The main part of a knowledge graph is a collection of triples, with each triple $ (h, t, r) $ denoting the fact that relation $ r $ exists between head entity $ h $ and tail entity $ t $. This can also be formalized as a labeled directed multigraph where each triple $ (h, t, r) $ represents a directed edge from node $ h $ to node $ t $ with label $ r $. Therefore, it is straightforward to build knowledge graphs for scholarly data by representing natural connections between scholarly entities with triples such as \textit{(AuthorA, Paper1, write)} and \textit{(Paper1, Paper2, cite)}.

Notably, instead of using knowledge graphs directly in some tasks, we can model them by knowledge graph embedding methods, which represent entities and relations as embedding vectors in semantic space, then model the interactions between them to solve the knowledge graph completion task. There are many approaches \cite{tran_analyzingknowledgegraph_2019} to modeling the interactions between embedding vectors resulting in many knowledge graph embedding methods such as ComplEx \cite{trouillon_complexembeddingssimple_2016} and CP$ _h $ \cite{lacroix_canonicaltensordecomposition_2018}. In the case of word embedding methods such as word2vec, embedding vectors are known to contain rich semantic information that enables them to be used in many semantic applications \cite{mikolov_distributedrepresentationswords_2013}. However, the semantic structures in the knowledge graph embedding space are not well-studied, thus knowledge graph embeddings are only used for knowledge graph completion but remain absent in the toolbox for data analysis of heterogeneous data in general and scholarly data in particular, although they have the potential to be highly effective and efficient. In this paper, we address these issues by providing a theoretical understanding of their semantic structures and designing a general semantic query framework to support data exploration.

For theoretical analysis, we first analyze the state-of-the-art knowledge graph embedding model CP$ _h $ \cite{lacroix_canonicaltensordecomposition_2018} in comparison to the popular word embedding model word2vec skipgram \cite{mikolov_distributedrepresentationswords_2013} to explain its components and provide understandings to its semantic structures. We then define the \textit{semantic queries on the knowledge graph embedding spaces}, which are algebraic operations between the embedding vectors in the knowledge graph embedding space to solve queries such as similarity and analogy between the entities on the original datasets. 

Based on our theoretical results, we design a general framework for data exploration on scholarly data by semantic queries on knowledge graph embedding space. The main component in this framework is the conversion between the data exploration tasks and the semantic queries. We first outline the semantic query solutions to some traditional data exploration tasks, such as similar paper prediction and similar author prediction. We then propose a group of new interesting tasks, such as analogy query and analogy browsing, and discuss how they can be used in modern digital libraries.

%
%


\section{Related Work} \label{sect:relatedwork}
\subsection{Knowledge graph for scholarly data}
Knowledge graph has gradually become the standard data format for heterogeneous and complicated datasets \cite{ehrlinger_definitionknowledgegraphs_2016}. There have been several attempts to build knowledge graph for scholarly data, either adopting the scholarly network directly \cite{wang_acekglargescaleknowledge_2018}, or deriving the knowledge graph from some similarity measures \cite{vahdati_unveilingscholarlycommunities_2018} \cite{sadeghi_integrationscholarlycommunication_2017}, or constructing the knowledge graph from survey papers \cite{fathalla_knowledgegraphrepresenting_2017}. However, they mostly focus on the data format or graph inference aspects of knowledge graph. In this paper, we instead focus on the knowledge graph embedding methods and especially the application of embedding vectors in data exploration.

\subsection{Knowledge graph embedding}

For a more in depth survey of knowledge graph embedding methods, please refer to \cite{tran_analyzingknowledgegraph_2019}, which defines their architecture, categorization, and interaction mechanisms. In this paper, we only focus on the semantic structures of the state-of-the-art model CP$ _h $ \cite{lacroix_canonicaltensordecomposition_2018}, which is an extension of CP \cite{hitchcock_expressiontensorpolyadic_1927}. 

In CP, each entity $ e $ has two embedding vectors $ \ve $ and $ \ve^{(2)} $ depending on its role in a triple as head or as tail, respectively. CP$ _h $ augments the data by making an inverse triple $ (t, h, r^{(a)}) $ for each existing triple $ (h, t, r) $, where $ r^{(a)} $ is the augmented relation corresponding to $ r $. When maximizing the likelihood by stochastic gradient descent, its score function is the sum:
\begin{equation} \label{eq:cph}
\begin{split}
\gS(h, t, r) =\ &\langle \vh, \vt^{(2)}, \vr \rangle + \langle \vt, \vh^{(2)}, \vr^{(a)} \rangle,
\end{split}
\end{equation}
where $ \vh, \vh^{(2)}, \vt, \vt^{(2)}, \vr, \vr^{(a)} \in \sR^{D} $ are the embedding vectors of $ h $, $ t $, and $ r $, respectively, and the trilinear-product $ \langle \cdot, \cdot, \cdot \rangle $ is defined as:
\begin{equation} \label{eq:trilinear}
\begin{split}
\langle \vh, \vt, \vr \rangle =\ &\sum_{d=1}^{D} h_d t_d r_d,
\end{split}
\end{equation}
where $ D $ is the embedding size and $ d $ is the dimension for which $ h_d $, $ t_d $, and $ r_d $ are the scalar entries.

The validity of each triple is modeled as a Bernoulli distribution and its validity probability is computed by the standard logistic function $ \sigma(\cdot) $ as:
\begin{equation} \label{eq:bernoulligeneral}
\begin{split}
P(1 | h, t, r) = \sigma(\gS(h, t, r)).
\end{split}
\end{equation}

\subsection{Word embedding}
The most popular word embedding models in recent years are word2vec variants such as word2vec skipgram \cite{mikolov_distributedrepresentationswords_2013}, which predicts the context-words $ c_i $ independently given the target-word $ w $, that is:
\begin{equation} \label{eq:langmodelskipgram}
\begin{split}
&P(c_i | w), \text{ where } i = 1, \dots, m.
\end{split}
\end{equation}

In practice, the expensive softmax functions in these multinoulli distributions are avoided by approximating them with negative sampling and solve for the Bernoulli distributions by using the standard logistic function $ \sigma(\cdot) $:
\begin{equation} \label{eq:skipgram}
\begin{split}
&P(1 | c_i, w) = \sigma (\vu_{c_i}^\top \vv_w), \text{ where } i = 1, \dots, m,
\end{split}
\end{equation}
where $ \vu_{c_i} $ is the context-embedding vector of context-word $ c_i $ and $ \vv_w $ is the word-embedding vector of target-word $ w $.


\section{Theoretical analysis}  \label{sect:theory}
Word2vec skipgram and its semantic structures are well-studied both theoretically and empirically \cite{mikolov_distributedrepresentationswords_2013}. CP$ _h $ is a new state of the art among many knowledge graph embedding models. We first ground the theoretical basis of CP$ _h $ on word2vec skipgram to explain its components and understand its semantic structures. We then define semantic queries on knowledge graph embedding space.

\subsection{The semantic structures of CP$ _h $} \label{sect:connection}
We first look at Eq. \ref{eq:skipgram} of word2vec skipgram and consider only one context-word $ c $ for simplicity. We can write the probability in proportional format as:
\begin{equation} \label{eq:skipgrampropto}
\begin{split}
P(1 | c, w) \propto\ &\exp \left( \vu_c^\top \vv_w \right).
\end{split}
\end{equation}

Note that the context-word $ c $ and target-word $ w $ are ordered and in word2vec skipgram, the target-word is the central word in a sliding window, e.g., $ w_i $ is the target-word and $ w_{i-k}, \dots, w_{i-1}, w_{i+1}, \dots, w_{i+k} $ are context-words. Therefore, the roles in each word pair are symmetric over the whole dataset. When maximizing the likelihood by stochastic gradient descent, we can write the approximate probability of unordered word pair and expand the dot products as:
\begin{align}
P(1 | c, w; w, c) \propto\ &\exp \left( \vu_c^\top \vv_w + \vu_w^\top \vv_c \right) \label{eq:skipgramproptounordered} \\
\propto\ &\exp \left( \sum_{d=1}^{D} {u_c}_d {v_w}_d + \sum_{d=1}^{D} {u_w}_d {v_c}_d \right), \label{eq:skipgramproptounorderedexpand}
\end{align}
where $ \vu_c $ and $ \vv_c $ are the context-embedding and word-embedding vectors of $ c $, respectively, $ \vu_w $ and $ \vv_w $ are the context-embedding and word-embedding vectors of $ w $, respectively, and $ {u_c}_d, {v_c}_d, {u_w}_d $, and $ {v_w}_d $ are their scalar entries, respectively.

We now return to Eq. \ref{eq:cph} of CP$ _h $ to also write the probability in Eq. \ref{eq:bernoulligeneral} in proportional format and expand the trilinear products according to Eq. \ref{eq:trilinear} as:
\begin{align}
P(1 | h, t, r) \propto\ &\exp \left( \langle \vh, \vt^{(2)}, \vr \rangle + \langle \vt, \vh^{(2)}, \vr^{(a)} \rangle \right) \label{eq:cphpropto} \\
\propto\ &\exp \left( \sum_{d=1}^{D} h_d t^{(2)}_d r_d + \sum_{d=1}^{D} t_d h^{(2)}_d r^{(a)}_d \right), \label{eq:cphproptoexpand}
\end{align}
where $ \vh, \vh^{(2)} $, $ \vt, \vt^{(2)} $, $ \vr, \vr^{(a)} $ are knowledge graph embedding vectors and $ h_d, h^{(2)}_d $, $ t_d, t^{(2)}_d $, $ r_d, r^{(a)}_d $ are the scalar entries.

Comparing Eq. \ref{eq:skipgramproptounorderedexpand} of word2vec skipgram and Eq. \ref{eq:cphproptoexpand} of CP$ _h $, we can see they have essentially the same form and mechanism. Note that the embedding vectors in word2vec skipgram are learned by aligning each target-word to different context-words and vice versa, which is essentially the same for CP$ _h $ by aligning each head entity to different tail entities in different triples and vice versa, with regards to the dimensions weighted by each relation. This result suggests that the \textit{semantic structures} of CP$ _h $ are similar to those in word2vec skipgram and we can use the head-role-based entity embedding vectors, such as $ \ve $, for semantic applications similarly to word embedding vectors. The tail-role-based entity embedding vectors, such as $ \ve^{(2)} $, contain almost the same information due to their symmetric roles, thus can be discarded in semantic tasks, which justifies this common practices in word embedding applications \cite{mikolov_distributedrepresentationswords_2013}.

\subsection{Semantic query} \label{sect:semquery}
We mainly concern with the two following structures of the embedding space.
\begin{itemize}
	\item \textit{Semantic similarity structure:} Semantically similar entities are close to each other in the embedding space, and vice versa. This structure can be identified by a vector similarity measure, such as the dot product between two embedding vectors. The similarity between two embedding vectors is computed as:
	\begin{equation} \label{eq:semsimilar}
	\begin{split}
	sim(\ve_1, \ve_2) = \ve_1^\top \ve_2.
	\end{split}
	\end{equation}
	
	\item \textit{Semantic direction structure:} There exist semantic directions in the embedding space, by which only one semantic aspect changes while all other aspects stay the same. It can be identified by a vector difference, such as the subtraction between two embedding vectors. The semantic direction between two embedding vectors is computed as:
	\begin{equation} \label{eq:semdirection}
	\begin{split}
	dir(\ve_1, \ve_2) = \ve_1 - \ve_2.
	\end{split}
	\end{equation}	
\end{itemize}

The algebraic operations, which include the above dot product and vector subtraction, or their combinations, can be used to approximate some important tasks on the original data. To do this, we first need to convert the data exploration task to the appropriate operations. We then conduct the operations on the embedding vectors and obtain the results. This process is defined as following.

\begin{defn}{\textit{Semantic queries on knowledge graph embedding space}} are defined as the algebraic operations between the knowledge graph embedding vectors to approximate a given data exploration task on the original dataset.
\end{defn}

\section{Semantic query framework} \label{sect:framework}
\begin{figure}[ht]
	\centering
	\includegraphics[width=1\linewidth]{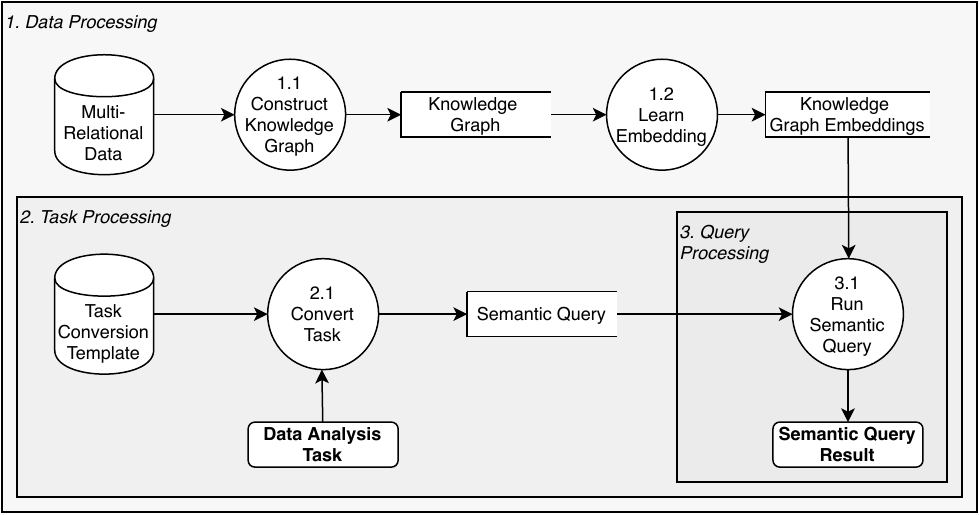}
	\caption{Architecture of the semantic query framework, with three main components Data Processing, Task Processing, and Query Processing. Notations follow Yourdon and Coad's data flow diagram convention, with circle denoting process, cylinder denoting database, open rectangle denoting data store, rectangle denoting input and output.}
	\label{fig:framework}
\end{figure}

Given the theoretical results, here we design a general framework for scholarly data exploration by using semantic queries on knowledge graph embedding space. Figure \ref{fig:framework} shows the architecture of the proposed framework. There are three main components, namely \textit{data processing}, \textit{task processing}, and \textit{query processing}.

\noindent\textbf{Data processing:} with two steps, (1) \textit{constructing the knowledge graph from scholarly data} by using the scholarly graph directly with entities such as \textit{authors}, \textit{papers}, \textit{venues}, and relations such as \textit{author-write-paper}, \textit{paper-cite-paper}, \textit{paper-in-venue}, and (2) \textit{learning the knowledge graph embeddings} as in \cite{tran_analyzingknowledgegraph_2019}. 

\noindent\textbf{Task processing:} converting data exploration tasks to algebraic operations on the embedding space by following task-specific conversion templates. Some important tasks and their conversion templates are discussed in Section \ref{sect:newtask}.

\noindent\textbf{Query processing:} executing semantic query on the embedding space and return results. Note that the algebraic operations on embedding vectors are linear and can be performed in parallel. Therefore, the \textit{semantic query} is efficient.

Note that the proposed \textit{semantic query framework} makes no assumption on the specific knowledge graph embedding models and the induced embedding spaces. Any embedding space that contains rich semantic information such as the listed \textit{semantic structures} can be applied in this framework.

\section{Exploration tasks and semantic queries conversion} \label{sect:newtask}
Here we present and discuss the semantic queries for some traditional and newly proposed scholarly data exploration tasks. The extended semantic query method is presented in the Ph.D. thesis \cite{tran_multirelationalembeddingknowledge_2020}.

\subsection{Similar entities}
\noindent\textbf{Tasks} Given an entity $ e \in \gE $, find entities that are similar to $ e $. For example, given \textit{AuthorA}, find authors, papers, and venues that are similar to \textit{AuthorA}. Note that we can restrict to find specific entity types. This is a traditional tasks in scholarly data exploration, whereas other below tasks are new. 

\noindent\textbf{Semantic query} We can solve this task by looking for the entities with highest similarity to $ e $. For example, the first result is:
\begin{equation} \label{eq:tasksimilarentity}
\begin{split}
\text{Result} =\ &\argmax_{e_i \in \gE} sim\left( \ve_i, \ve \right).
\end{split}
\end{equation}

\subsection{Similar entities with bias}
\noindent\textbf{Tasks} Given an entity $ e \in \gE $ and some positive bias entities $ A = \{a_1, \dots, a_k\} $ known as expected results, find entities that are similar to $ e $ following the bias in $ A $. For example, given \textit{AuthorA} and some successfully collaborating authors, find other similar authors that may also result in good collaborations with \textit{AuthorA}. 

\noindent\textbf{Semantic query} We can solve this task by looking for the entities with highest similarity to both $ e $ and $ A $. For example, denoting the arithmetic mean of embedding vectors in $ A $ as $ \bar{A} $, the first result is:
\begin{equation} \label{eq:tasksimilarentitybias}
\begin{split}
\text{Result} =\ &\argmax_{e_i \in \gE} sim \left( \ve_i, \bar{A} + \ve \right).
\end{split}
\end{equation}

\subsection{Analogy query}
\noindent\textbf{Tasks} Given an entity $ e \in \gE $, positive bias $ A = \{a_1, \dots, a_k\} $, and negative bias $ B = \{b_1, \dots, b_k\} $, find entities that are similar to $ e $ following the biases in $ A $ and $ B $. The essence of this task is tracing along a semantic direction defined by the positive and negative biases. For example, start with \textit{AuthorA}, we can trace along the \textit{expertise direction} to find authors that are similar to \textit{AuthorA} but with higher or lower expertise.

\noindent\textbf{Semantic query} We can solve this task by looking for the entities with highest similarity to $ e $ and $ A $ but not $ B $. For example, denoting the arithmetic mean of embedding vectors in $ A $ and $ B $ as $ \bar{A} $ and $ \bar{B} $, respectively, note that $ \bar{A} - \bar{B} $ defines the semantic direction along the positive and negative biases, the first result is:
\begin{equation} \label{eq:taskanalogyquery}
\begin{split}
\text{Result} =\ &\argmax_{e_i \in \gE} sim \left( \ve_i, \bar{A} - \bar{B} + \ve \right).
\end{split}
\end{equation}

\subsection{Analogy browsing}
\noindent\textbf{Tasks} This task is an extension of the above analogy query task, by tracing along multiple semantic directions defined by multiple pairs of positive and negative biases. This task can be implemented as an interactive data analysis tool. For example, start with \textit{AuthorA}, we can trace to authors with higher expertise, then continue tracing to new domains to find all authors similar to \textit{AuthorA} with high expertise in the new domain. For another example, start with \textit{Paper1}, we can trace to papers with higher quality, then continue tracing to new domain to look for papers similar to \textit{Paper1} with high quality in the new domain.

\noindent\textbf{Semantic query} We can solve this task by simply repeating the semantic query for analogy query with each pair of positive and negative bias. Note that we can also combine different operations in different order to support flexible browsing.

\section{Conclusion} \label{sect:conclusion}
In this paper, we studied the application of knowledge graph embedding in exploratory data analysis. We analyzed the CP$ _h $ model and provided understandings to its semantic structures. We then defined the \textit{semantic queries on knowledge graph embedding space} to efficiently approximate some operations on heterogeneous data such as scholarly data. We designed a general framework to systematically apply semantic queries to solve scholarly data exploration tasks. Finally, we outlined and discussed the solutions to some traditional and pioneering exploration tasks emerged from the semantic structures of the knowledge graph embedding space.

This paper is dedicated to the theoretical foundation of a new approach and discussions of emerging tasks, whereas experiments and evaluations are left for the future work. There are several other promising directions for future research. One direction is to explore new tasks or new solutions of traditional tasks using the proposed method. Another direction is to implement the proposed exploration tasks on real-life digital libraries for online evaluation.

\section*{Acknowledgments}
This work was supported by “Cross-ministerial Strategic Innovation Promotion Program (SIP) Second Phase, Big-data and AI-enabled Cyberspace Technologies” by New Energy and Industrial Technology Development Organization (NEDO).

\begin{spacing}{1.0}
\bibliographystyle{splncs04}

\end{spacing}

\clearpage
\appendix

\section{KG20C: scholarly knowledge graph benchmark dataset}
Some methods described in this paper can be evaluated on scholarly knowledge graphs. We recommend using the KG20C benchmark dataset, available at \url{https://github.com/tranhungnghiep/KG20C}. Details of the dataset will be officially presented in a forthcoming peer-reviewed publication.

\end{document}